%% file: main.tex
\documentclass[]{ceurart} 
\usepackage{listings}
\usepackage{cleveref}
\input{math_commands}

\input{commands}
\usepackage{subcaption}
\usepackage{wrapfig}
\usepackage{placeins}

\crefname{equation}{Eq.}{Eqs.}
\crefname{figure}{Fig.}{Figs.}
\crefname{table}{Tab.}{Tabs.}
\crefname{section}{Sec.}{Secs.}

\begin{document}

\copyrightyear{2026} 
\copyrightclause{Copyright for this paper by its authors.
  Use permitted under Creative Commons License Attribution 4.0
  International (CC BY 4.0).}
\conference{Preprint} 

\title{Building Trust in PINNs: Error Estimation through Finite Difference Methods}


\author[1]{Aleksander Krasowski}[
email=aleksander.krasowski@hhi.fraunhofer.de,
]
\cormark[1]
\address[1]{Department of Artificial Intelligence, Fraunhofer Heinrich Hertz Institute}
\address[2]{Department of Electrical Engineering and Computer Science, Technische Universität Berlin}
\address[3]{BIFOLD - Berlin Institute for the Foundations of Learning and Data}
\address[4]{Centre of eXplainable Artificial Intelligence, Technological University Dublin}

\author[1]{René P. Klausen}
\author[1]{Aycan Celik}
\author[1,4]{Sebastian Lapuschkin}
\author[1,2,3]{Wojciech Samek}
\author[1]{Jonas Naujoks}[
email=jonas.naujoks@hhi.fraunhofer.de
]
\cormark[1]

\cortext[1]{Corresponding author.}

\begin{abstract}
  Physics-informed neural networks (PINNs) constitute a flexible deep learning approach for solving partial differential equations (PDEs), which model phenomena ranging from heat conduction to quantum mechanical systems. 
  Despite their flexibility, PINNs offer limited insight into how their predictions deviate from the true solution, hindering trust in their prediction quality.
  We propose a lightweight post-hoc method that addresses this gap by producing pointwise error estimates for PINN predictions, which offer a natural form of explanation for such models, identifying not just whether a prediction is wrong, but where and by how much. 
  For linear partial differential equations, the error between a 
  PINN approximation and the true solution satisfies the same differential operator as the original problem, but driven by the PINN's PDE residual as its source term.
  We solve this error equation numerically using finite difference methods requiring no knowledge of the true solution. 
  Evaluated on several benchmark PDEs, our method yields accurate error maps at low computational cost, enabling targeted and interpretable validation of PINNs.
\end{abstract}


\maketitle

\section{Introduction}
A central aim in the physical sciences is the prediction and modeling of phenomena that arise in the natural world. Among the most fundamental tools for this are partial differential equations (PDEs). These equations describe how quantities evolve over space and time. For example, possible applications include the motion of planetary objects, the conduction of heat through a solid, the spread of a disease through a population, and the behavior of quantum mechanical systems. A solution to such an equation then provides a complete quantitative description of the phenomenon in question. However, closed-form solutions are available for only the simplest cases, and one must resort to numerical approximations.
With the rise of artificial intelligence and ever-growing computational capabilities, neural networks have been explored as an alternative to classical numerical methods.
Physics-informed neural networks (PINNs) \cite{RaissiPhysicsinformedNeuralNetworks2019} are one such approach
for solving PDEs by incorporating the governing laws as prior physical knowledge into the training objective. 
Since their introduction, PINNs have been successfully applied across a wide range of domains \cite{karniadakisPhysicsinformedMachineLearning2021,cuomoScientificMachineLearning2022,ToscanoPINNsPIKANsRecent2024}.

Despite their flexibility, they are far from flawless.
A growing body of research has documented systematic pitfalls arising from their composite loss formulation \cite{WangUnderstandingMitigatingGradient2021,WangWhenWhyPINNs2022,RathoreChallengesTrainingPINNs2024}, 
as well as problems in propagating the solution from the initial condition throughout the domain \cite{KrishnapriyanCharacterizingPossibleFailure2021,DawMitigatingPropagationFailures2023}.
PINNs may hence produce incorrect solutions that are difficult to identify from the training loss alone, 
limiting their reliability and trustworthiness.
Reliable deployment of such models therefore requires methods that can both identify where predictions fail and by how much, ideally without access to the true solution.
\newpage

At the same time, interpretability has become a central concern in scientific machine learning \cite{WetzelInterpretableMachineLearning2025}, where trust in model predictions is grounded in physical consistency rather than statistical fits alone. For PINNs, such an explainable artificial intelligence (XAI) perspective remains largely absent.
New models call for novel forms of explanation. In established XAI domains such as computer vision, explanations typically identify which input features drove a prediction. This paradigm does not transfer to PINNs, where inputs are spatio-temporal coordinates rather than semantically meaningful features. 
On the other hand, PINNs offer a benefit rarely available in other machine learning settings: the governing equation that the ground truth solution must satisfy is known by construction and already embedded in the training objective.
While the resulting residual quantifies local PDE violations, it does not directly reveal how far the approximation deviates from the actual ground truth solution.
A method that efficiently converts these residuals into error estimates would therefore provide a natural and actionable form of explanation, analogous to attribution heat maps in computer vision, but quantifying where and by how much a prediction is wrong.
In this work, we propose precisely such a method: a lightweight post-hoc approach that yields spatially and temporally resolved error maps for PINN predictions to efficiently validate and interpret model behavior.
This quantitative angle on model reliability for PINNs enables practitioners to trust or reject predictions at specific locations in the domain.

We focus on linear PDEs and exploit a key property: for a PINN approximation of a linear PDE, the pointwise error between the prediction and the true solution satisfies the same differential operator as the original problem, driven by the PINN's PDE residual.
We solve this error equation via finite difference methods, without knowledge of the true solution, and evaluate on five benchmark settings, demonstrating accurate error maps at low computational cost.

\paragraph{Related Work}

Prior work on PINN reliability includes both a priori guarantees 
\cite{DeRyckGenericBounds2022,MishraEstimatesGeneralizationError2023,RyckNumericalAnalysisPhysicsinformed2024}  and a posteriori bounds \cite{EirasEfficientErrorCertification2024,HillebrechtCertifiedMachineLearning2022,HillebrechtPredictionErrorCertification2025,ErnstPosterioriCertificationNeural2025}. \cite{HillebrechtCertifiedMachineLearning2022,HillebrechtRigorousPosterioriError2025} derive rigorous error bounds for PINNs, based on constants derived from the strongly continuous semigroup of the governing PDE operator. Recently, \cite{HillebrechtPredictionErrorCertification2025} extended this approach by providing a method to derive these constants using finite differences. Crucially, however, finite differences are used there only to parametrize the bound, not to integrate the error equation itself.
We note that unlike \cite{HillebrechtRigorousPosterioriError2025,LiuResidualbasedErrorBound2023}, our estimates do not constitute provable upper bounds; however, we empirically show that they accurately track the true error efficiently.

Closely related to our work, \cite{LiuResidualbasedErrorBound2023} and \cite{FanPhysicsInformedInferenceTime2025} also build on the same error equation driven by the PINN's PDE residual.
\cite{LiuResidualbasedErrorBound2023} invert the defect equation to derive provable error bounds, though their approach is limited to ODEs and first-order PDEs.
\cite{FanPhysicsInformedInferenceTime2025}, on the other hand, solve the error equation stochastically via multilevel Monte Carlo to correct PINN solutions in high-dimensional settings.
In contrast, we consider low-dimensional PDEs and directly integrate the defect equation numerically using finite differences, yielding cheap, pointwise error estimates, applicable for a broad class of PDEs.

\section{Theoretical Background}

\subsection{Physics-Informed Neural Networks}
\label{sec:PINNs}

PINNs \cite{RaissiPhysicsinformedNeuralNetworks2019} are a deep learning approach aimed at solving initial-boundary value problems (IBVPs) by incorporating the governing equations as constraints during training. Let $\Omega\subset \mathbb R^n$ denote an open, bounded, connected domain, and $\Gamma_k\subseteq\partial\Omega$ be smooth components of the boundary,
where $x \in \Omega$ may be purely spatial or may include time.
Then we define the IBVP by
\begin{align}
    & \mathcal{D}[u](x) = 0, \quad x \in \Omega \label{eq:IBVPpde}\\
    & \mathcal B_k[u](x) = 0 , \quad x \in \Gamma_k \ \mathrm{for}\ k=1,\ldots,K\label{eq:IBC} \quad\mathrm{.}
\end{align}
where $\mathcal{D}$ is a differential operator encoding the governing PDE and $\mathcal{B}$ represents the initial-boundary condition (IBC) at $\Gamma_k$, where $k=1,\ldots,K$. We say that $u:\Omega \to \mathbb R^d$ is a solution to the IBVP defined by $\mathcal D, \mathcal B_1,\ldots, \mathcal B_K$, if it satisfies \cref{eq:IBVPpde,eq:IBC}. Note that \cref{eq:IBC} is formulated to capture both Dirichlet and Neumann conditions, among others.

To find such a solution, PINNs approximate it by a neural network $\varphi(x; \theta)$, which depends on parameters $\theta\in\Theta$. Training proceeds by minimizing a composite loss that penalizes violations of both the governing PDE and the prescribed boundary conditions 
\begin{align}
    \label{eq:physics_loss}
    \mathcal L (\theta) = \frac{1}{|\mathcal X_\mathrm{pde}|} \sum_{x\in\mathcal X_\mathrm{pde}} \norm{\mathcal{D}[\varphi(x; \theta)]}_2^2 + \sum_{k=1}^K \frac{1}{|\mathcal X_\mathrm{bc,k}|} \sum_{x\in\mathcal X_\mathrm{bc,k}} \norm{\mathcal B_k [\varphi(x; \theta)]}_2^2\quad,
\end{align}
over sampled training points $\mathcal X_\mathrm{pde}\subset\Omega$ and $\mathcal X_\mathrm{bc,k}\subset\Gamma_k$.

A common approach to mitigating the optimization difficulties inherent to this composite loss structure is hard-constraining several or all IBCs such that they are fulfilled by design \cite{WangExpertGuideTraining2023}.
Given an IBVP with initial condition $u(x,t_0)=u_0(x)$ and Dirichlet boundary conditions $u(x_L,t)=u(x_R,t)=0$, we can transform the PINN as
\begin{align}
    \varphi (x,t;\theta) = u_0(x) + (t-t_0) (x-x_L)(x-x_R) \phi(x, t; \theta)\quad, \label{eq:hard_constraining}
\end{align}
which satisfies these conditions for any PINN $\phi$.
Throughout this work, we hard-constrain all IBCs, ensuring that the PDE residual is the sole source of the approximation error.

\subsection{Finite Difference Methods}
\label{sec:fdm}

Unlike most machine learning models, PINNs solve problems with known PDEs, which allows for generating reference solutions via numerical methods. 
In the following, we introduce finite difference methods (FDM) \cite{StrikwerdaFiniteDifferenceSchemes2004}, a widely used numerical scheme for approximating solutions to IBVPs, and show
in \cref{sec:methodology} how this framework can be exploited to achieve post-hoc error diagnostics directly.

The core idea is to discretize the domain $\Omega$ into a grid and approximate derivatives with finite differences between neighboring grid points. 
For instance, given a (one-dimensional) grid with spacing $\Delta x$ and using the shorthand notation $u_i=u(x_i)$, the first and second spatial derivatives can be approximated as 
\begin{align}
    \frac{\partial u}{\partial x}\bigg|_{x_i} \approx \frac{u_{i+1} - u_{i-1}}{2\,\Delta x}\quad, \qquad
    \frac{\partial^2 u}{\partial x^2}\bigg|_{x_i} \approx \frac{u_{i+1} - 2u_i + u_{i-1}}{\Delta x^2}\quad.
    \label{eq:central_diff}
\end{align}
These so-called central differences are one of several possible finite difference schemes. 
Higher-order, mixed derivatives as well as alternative schemes follow analogously \cite{StrikwerdaFiniteDifferenceSchemes2004}. 

To illustrate the core idea, consider the one-dimensional Poisson equation on $\Omega = (x_L,x_R) \subset \mathbb{R}$ with homogeneous Dirichlet boundary conditions $u(x_L)=u(x_R)=0$, differential operator $\mathcal{D}[u] = \frac{\partial^2 u}{\partial x^2}$, and source term $f$, such that $\mathcal{D}[u] -f =0$.
We discretize the domain $\Omega$ into $k$ grid points $x_0,x_1,\dots,x_{k-1}$ with uniform spacing $\Delta x$, where $x_0=x_L$ and $x_{k-1}=x_R$. 
Approximating $\mathcal{D}$ at interior points using \cref{eq:central_diff} and assembling across all grid points together with the boundary conditions yields
\begin{align}
\small
    \frac{1}{\Delta x^2}
    \begin{bmatrix}
        \Delta x^2 & 0 & 0 & \cdots & 0 & 0 \\
        1  & -2 & 1 & \cdots & 0 & 0 \\
        0 & 1 & -2 & \cdots & 0 & 0 \\
        \vdots &  & \ddots & \ddots & & \vdots \\
        0 & 0 & \cdots & 1 & -2 & 1 \\
        0 & 0 & 0 & \cdots & 0 & \Delta x^2
    \end{bmatrix}
    \begin{bmatrix}
        u_0 \\ u_1 \\ u_2 \\ \vdots \\ u_{k-2} \\ u_{k-1}
    \end{bmatrix}
    -
    \begin{bmatrix}
        0 \\ f_1 \\ f_2 \\ \vdots \\ f_{k-2} \\ 0
    \end{bmatrix} = 0 \quad ,
    \label{eq:fdm_poisson_full}
\end{align}
We write this compactly as $\boldsymbol{L}\boldsymbol{u} - \boldsymbol{f}=0$, where $\boldsymbol{L}$ denotes the discretized spatial operator.
For other \emph{spatial} differential operators $\mathcal{D}$ the discretization $\boldsymbol{L}$ follows analogously. 

\paragraph{Time-Stepping} 
For time-dependent problems, we discretize the time interval $[0,t_\mathrm{max}]$ with spacing $\Delta t$ and write $u_{i}^{n} = u(x_i, t_n)$. 
Rather than solving the full spatio-temporal system at once, 
we exploit the causal structure and march forward in time from the initial condition $u^0$, solving a spatial system at each step. 
The choice of the time-stepping scheme depends on the order of the temporal derivative. 
Boundary conditions are incorporated analogously to \cref{eq:fdm_poisson_full} and omitted below for notational clarity.

\paragraph{First-order Time Derivatives (Crank-Nicolson)}
For first-order temporal derivatives, we employ the Crank-Nicolson scheme \cite{CrankPracticalMethodNumerical1947}, 
which averages the spatial operator between the current and next time level.
Taking the heat equation $\frac{\partial u}{\partial t}-\alpha \frac{\partial^2 u}{\partial x^2} = 0$ as an example, this yields
\begin{align}
    \frac{u_i^{n+1} - u_i^n}{\Delta t} - \frac{1}{2}\left[\frac{\alpha}{\Delta x^2} (u_{i-1}^{n+1} - 2u_i^{n+1} + u_{i+1}^{n+1}) + \frac{\alpha}{\Delta x^2} (u_{i-1}^n - 2u_i^n + u_{i+1}^n)\right] = 0\quad.
\end{align}
Note that this equation involves multiple unknowns at $t=n+1$, coupled across neighboring spatial points. 
This scheme is implicit, requiring solving a linear system at each time step. 
For any spatial operator $\boldsymbol{L}$ we can assemble the following system across all interior points
\begin{align}
    (\boldsymbol{I} - \tfrac{1}{2} \Delta t \boldsymbol{L})\boldsymbol{u}^{n+1} &= 
    (\boldsymbol{I} + \tfrac{1}{2} \Delta t \boldsymbol{L})\boldsymbol{u}^{n} \quad,
    \label{eq:crank_nicolson_matrix}
\end{align}
with $\boldsymbol{I}$ denoting the identity matrix and $\boldsymbol{u}^n = [u_1^n, \dots, u_{k-2}^n]$. 
Crank-Nicolson is second-order accurate in both space and time, and unconditionally stable for parabolic problems \cite{StrikwerdaFiniteDifferenceSchemes2004}.

\paragraph{Second-order Time Derivatives (Central Differences)}
For PDEs with second-order time derivatives, we apply central differences in time, mirroring the spatial stencil. 
In matrix form this becomes 
\begin{align}
    \boldsymbol{u}^{n+1} &= 2 \boldsymbol{u}^n + \Delta t^2 \cdot \boldsymbol{L} \boldsymbol{u}^n - \boldsymbol{u}^{n-1} \quad.
    \label{eq:second_order_time_central_diff}
\end{align}
Here the scheme only involves known variables on the right-hand side, making it explicit.
This allows for quick computations; however, introduces stability constraints \cite{StrikwerdaFiniteDifferenceSchemes2004}.

\section{Methodology}
\label{sec:methodology}
\blfootnote{Code is available at \url{https://github.com/aleks-krasowski/pinn-fdm-error-estimation}.}
Rather than using FDM solely to generate a reference solution, we show that for linear PDEs the same framework can be used to produce spatially resolved error maps that explain where and by how much a PINN prediction deviates from the true solution, directly from its PDE residual.

Let $\hat \varphi := \varphi(x;\hat\theta)$ denote a trained, hard-constrained PINN with parameters $\hat\theta$, which approximates the true solution $u(x)$ of a linear PDE $\mathcal{D}[u]
 = 0$.
The pointwise error of the PINN is 
    $e(x) := u(x) - \hat\varphi(x)$
and the PDE residual of the PINN is given by $R(x) = \mathcal{D}[\hat\varphi](x)$.
Given the linearity of the operator, applying it to the error yields the defect equation
\begin{align}
    \mathcal{D}[e](x) 
    &= \underbrace{\mathcal{D}[u](x)}_{=0} - \mathcal{D}[\hat\varphi](x) 
    = -R(x) \quad. 
\end{align}
Hence, the error $e$ satisfies the same PDE as the original problem, but with the PINN residual $R=\mathcal{D}[\hat \varphi]$ as a source term, which we can evaluate at any desired point $x \in \Omega$. 
Through hard-constraining, we make sure that the initial and boundary conditions are satisfied exactly, i.e., $e(x)=0$ for all initial and boundary points. 
This allows the recovery of the error $e$ from the PDE residual alone, without knowledge of the true solution.

\paragraph{Integrating the Residual using FDM}
We solve the error equation using the FDM framework as introduced in \cref{sec:fdm}. 
The discretization is structurally identical to that of the PDE. The spatial operator $\boldsymbol{L}$ and time-stepping remain the same.
Two things change, however: (i) we solve for the unknown error $\boldsymbol{e}$ instead of the solution $\boldsymbol{u}$, (ii) the PINN residual $\boldsymbol{R}$ appears as a source term.

For steady-state problems, the error equation becomes 
$\boldsymbol{L} \boldsymbol{e}  - \boldsymbol{R} = 0$,
where $\boldsymbol{R}$ is the PINN residual evaluated on the discrete grid points.
For time-dependent problems we march forward in time starting from $\boldsymbol{e}^0$, which is 0 due to hard-constraining.

For first-order time derivatives, the Crank-Nicolson update (\cref{eq:crank_nicolson_matrix}) yields
\begin{align}
    (\boldsymbol{I} - \tfrac{1}{2}\Delta t \boldsymbol{L})\boldsymbol{e}^{n+1} &= 
    (\boldsymbol{I} + \tfrac{1}{2}\Delta t \boldsymbol{L})\boldsymbol{e}^{n} - \tfrac{1}{2}\Delta t(\boldsymbol{R}^n + \boldsymbol{R}^{n+1})\quad.
\end{align}
Analogously, for problems with second-order time derivatives the central difference (\cref{eq:second_order_time_central_diff}) becomes
\begin{align}
    \boldsymbol{e}^{n+1} &= 2\boldsymbol{e}^n + \Delta t^2 \cdot (\boldsymbol{L} \boldsymbol{e}^n - \boldsymbol{R}^n) - \boldsymbol{e}^{n-1}\quad.
\end{align}

\section{Experiments \& Discussion}
We evaluate whether the proposed method produces error maps that accurately explain where and by how much a PINN prediction deviates from the true solution.
For this, we consider four benchmark PDEs (\cref{tab:problems}, \cref{fig:problems}), with the Poisson equation evaluated in both one and two dimensions.
All problems admit known analytical solutions $u$, enabling direct comparison of our error estimates against the ground truth.
\begin{table}[t]
    \centering
    \small
    \caption{Benchmark problems. 
    The initial condition is $u(x,0)=\sin(2 \pi x)$ for all time-dependent problems. Parameters are: $\alpha=\tfrac{1}{20}$, $\beta=2$, $c=\frac{1}{2}$. 
    }
    \label{tab:problems}
    \begin{tabular}{lll}
        \toprule
        Problem & PDE & Boundary Conditions\\
        \midrule
        Poisson & $\nabla^2u - f= 0$ & Dirichlet: $u(x_L)=u(x_R)=0$ \\ [0.3em]
        Heat & $\frac{\partial u}{\partial t} - \alpha \tfrac{\partial^2 u}{\partial x^2} = 0$ & Dirichlet: $u(x_L)=u(x_R)=0$ \\ [0.3em]
        Drift-Diffusion & $\tfrac{\partial u}{\partial t} - \alpha \tfrac{\partial^2 u}{\partial x^2} - \beta \tfrac{\partial u}{\partial x} = 0$ & Periodic: $u(x_L)=u(x_R)$ \\ [0.3em]
        Wave & $\tfrac{\partial^2 u}{\partial t^2} - c^2 \tfrac{\partial^2 u}{\partial x^2} = 0$ & Dirichlet: $u(x_L)=u(x_R)=0$ \\ 
        \bottomrule
    \end{tabular}
\end{table}
\newcommand{\problemfigwidth}{0.2\linewidth}
\begin{figure}
    \centering
    \begin{subfigure}[t]{\problemfigwidth}
        \includegraphics[width=\linewidth]{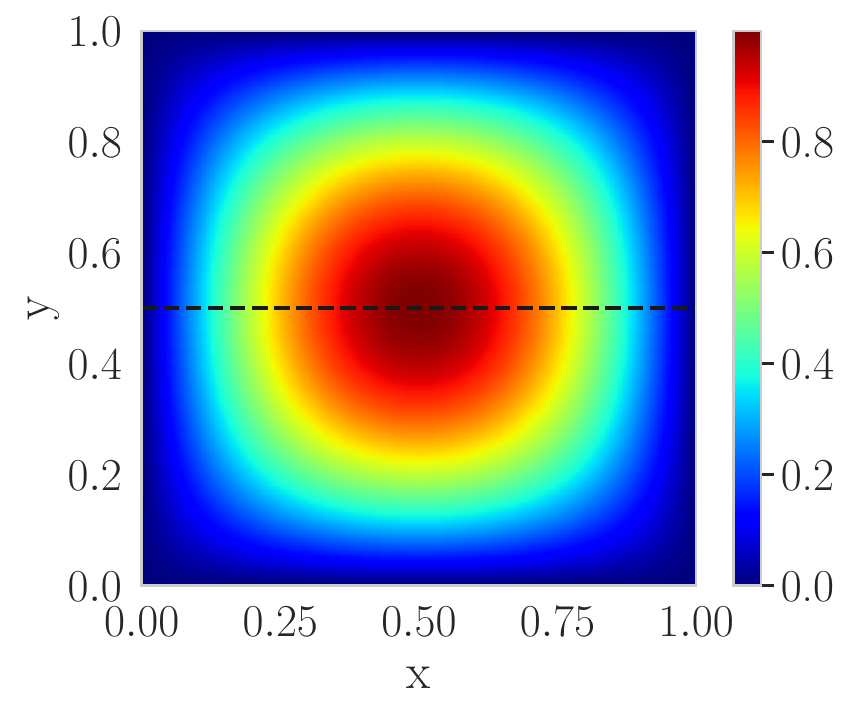}        
        \caption{Poisson 2D}
        \label{fig:poisson2d}
    \end{subfigure}
    \hfill 
    \begin{subfigure}[t]{\problemfigwidth}
        \includegraphics[width=\linewidth]{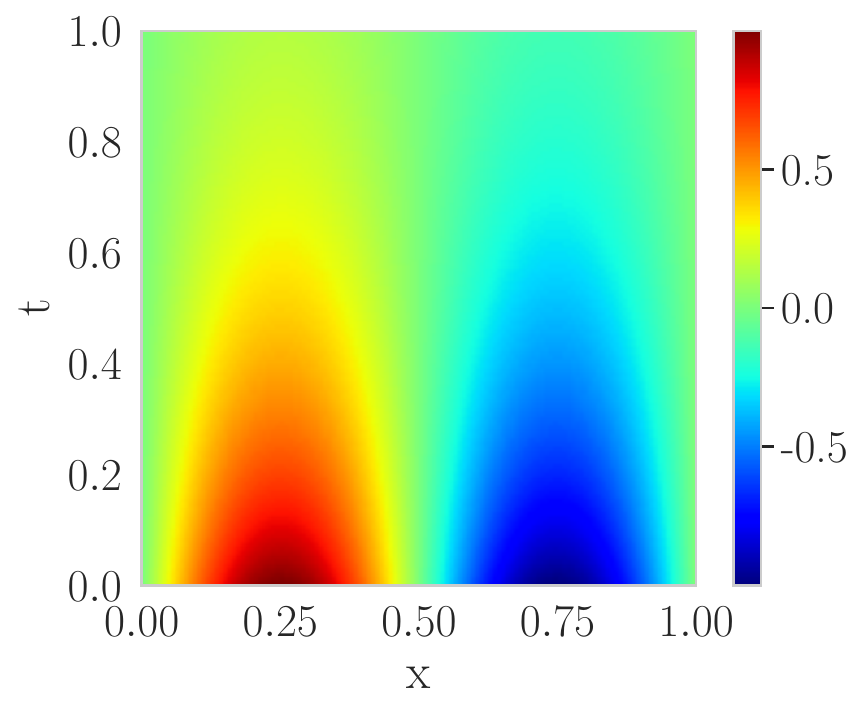}        
        \caption{Heat}
    \end{subfigure}
    \hfill 
    \begin{subfigure}[t]{\problemfigwidth}
        \includegraphics[width=\linewidth]{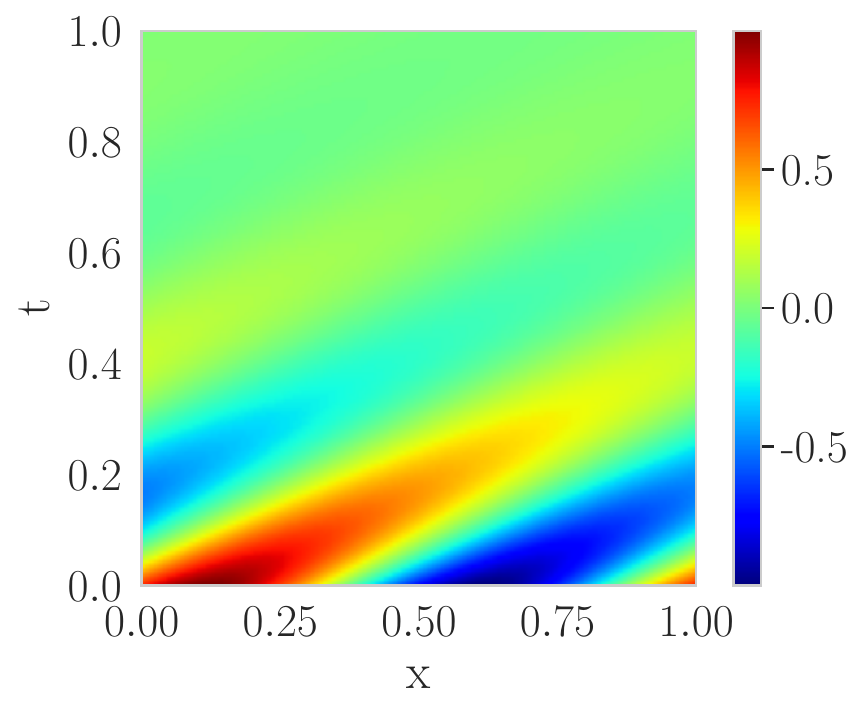}        
        \caption{Drift-Diffusion}
    \end{subfigure}
    \hfill 
    \begin{subfigure}[t]{\problemfigwidth}
        \includegraphics[width=\linewidth]{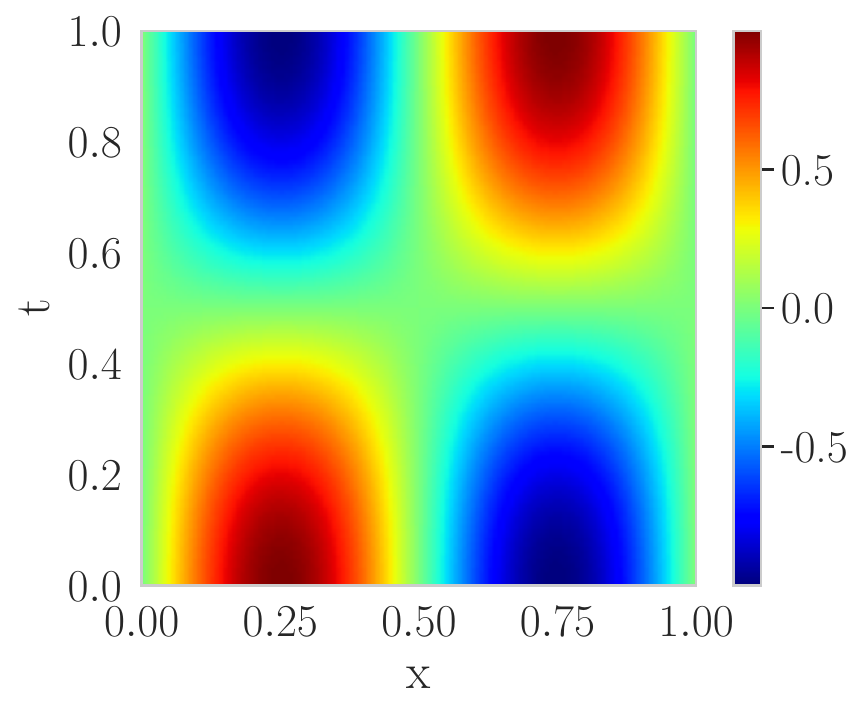}        
        \caption{Wave}
    \end{subfigure}
    \caption{Exact solutions for the benchmark problems. The horizontal line in \hyperref[fig:poisson2d]{(a)} at $y=0.5$ represents the solution for the 1D Poisson problem.}
    \label{fig:problems}
\end{figure}
For each problem we train a hard-constrained PINN $\varphi$ using the DeepXDE library \cite{LuDeepXDEDeepLearning2021}: an MLP with 3 hidden layers of 20 neurons, $\tanh$ activations, optimized with Adam \cite{KingmaAdamMethodStochastic2017} at a learning rate of $10^{-3}$.
We consider two PINN configurations, \emph{well-trained} ($10\,000$ collocation points and $10\,000$ iterations) and \emph{randomly initialized} (no training), to demonstrate that our method works across a range of PINN qualities.

We compare three error estimates against the true error $e_{\mathrm{true}} = u - \hat \varphi$:
\begin{enumerate}
    \item \textbf{Residual-based} (ours): $e_{\mathrm{res}}$, obtained by solving $\mathcal{D}[e] = -R$ via FDM (\cref{sec:methodology}).
    \item \textbf{FDM reference}: $e_{\mathrm{FDM}} = u_{\mathrm{FDM}} - \hat \varphi$, where $u_{\mathrm{FDM}}$ is obtained by solving the original PDE via FDM using the same grid.
    \item \textbf{Certified bound} \cite{HillebrechtRigorousPosterioriError2025}: $e_{\mathrm{bound}}$, a rigorous upper bound applied to the heat equation\footnote{The upper bound is given as $e_{\mathrm{bound}}(t) \leq \int_{s=0}^t Me^{\omega(t-s)} |\!|\widetilde R(\,\cdot\,,s)|\!|_2\, \mathrm{d}s$, with $M=1$ and $\omega=-\tfrac{1}{20}\pi^2$ derived from the semigroup of the PDE operator and $\widetilde R$ being a smoothed version of $R$, following \cite{HillebrechtRigorousPosterioriError2025}.}.
\end{enumerate}
$e_{\mathrm{FDM}}$ represents the standard approach for evaluating PINNs when analytical solutions are not available, i.e., comparing against a numerically obtained solution.
The certified bound $e_{\mathrm{bound}}$ represents a PINN-specific baseline proposed for error estimation \cite{HillebrechtRigorousPosterioriError2025}; however it offers no spatial resolution of the error and thus we only include it for comparison over the time domain.

\begin{figure}[t]
    \centering
    \fbox{\includegraphics[width=0.6\linewidth, trim=0 8pt 0 8pt, clip]{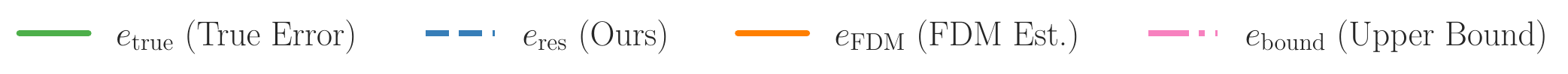}} \\ [0.5em]
    \begin{minipage}[c]{0.45\linewidth}
        \subcaptionbox{$t \approx 0.25$}{\includegraphics[width=0.48\linewidth]{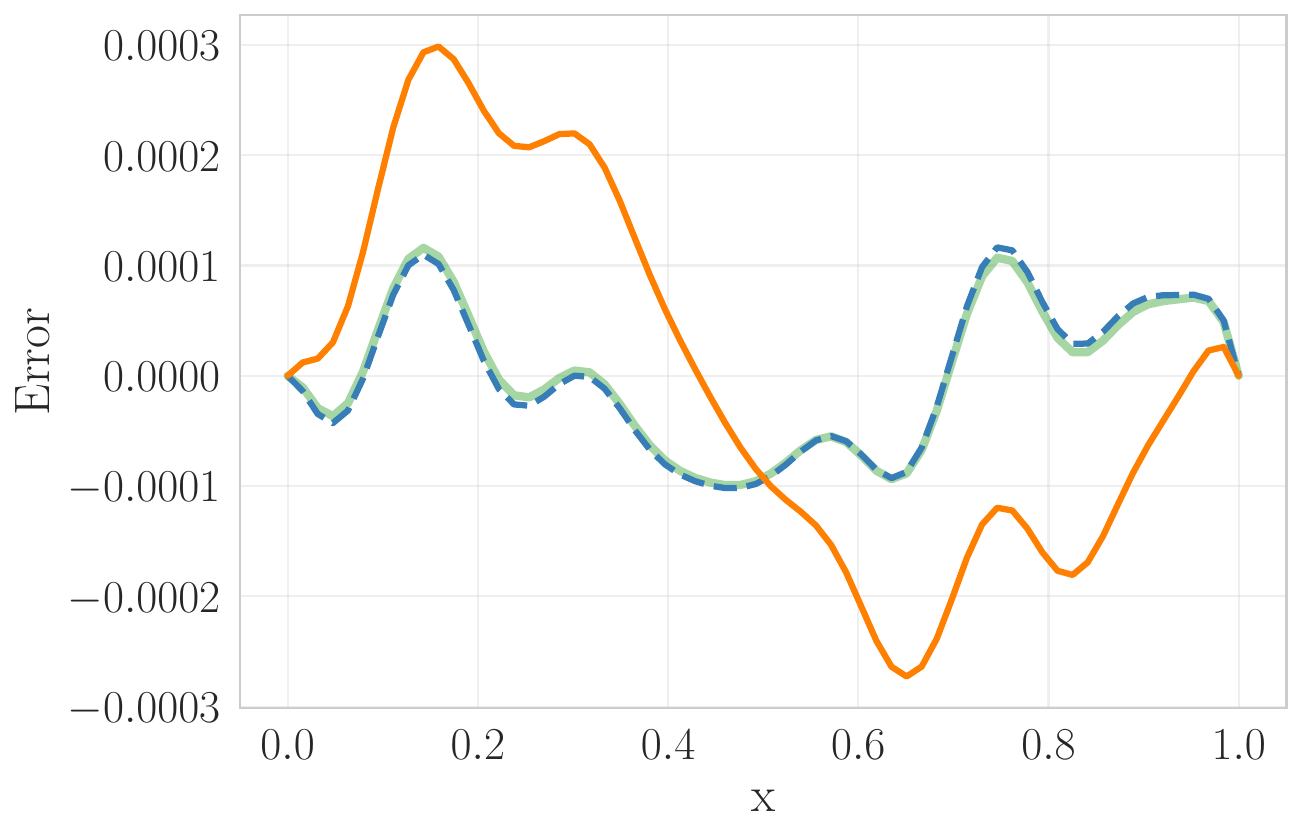}}\hfill
        \subcaptionbox{$t \approx 0.5$}{\includegraphics[width=0.48\linewidth]{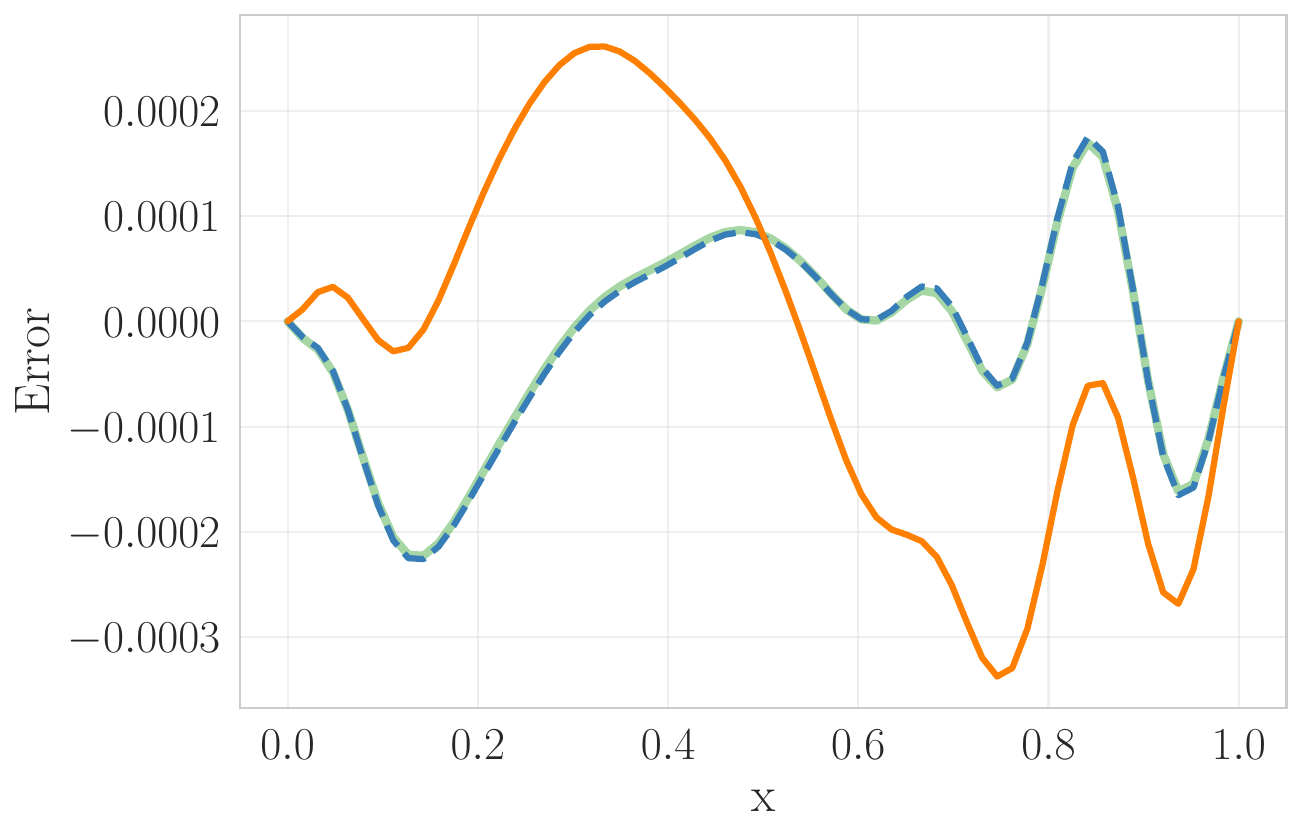}}\\[0.5em]
        \subcaptionbox{$t \approx 0.75$}{\includegraphics[width=0.48\linewidth]{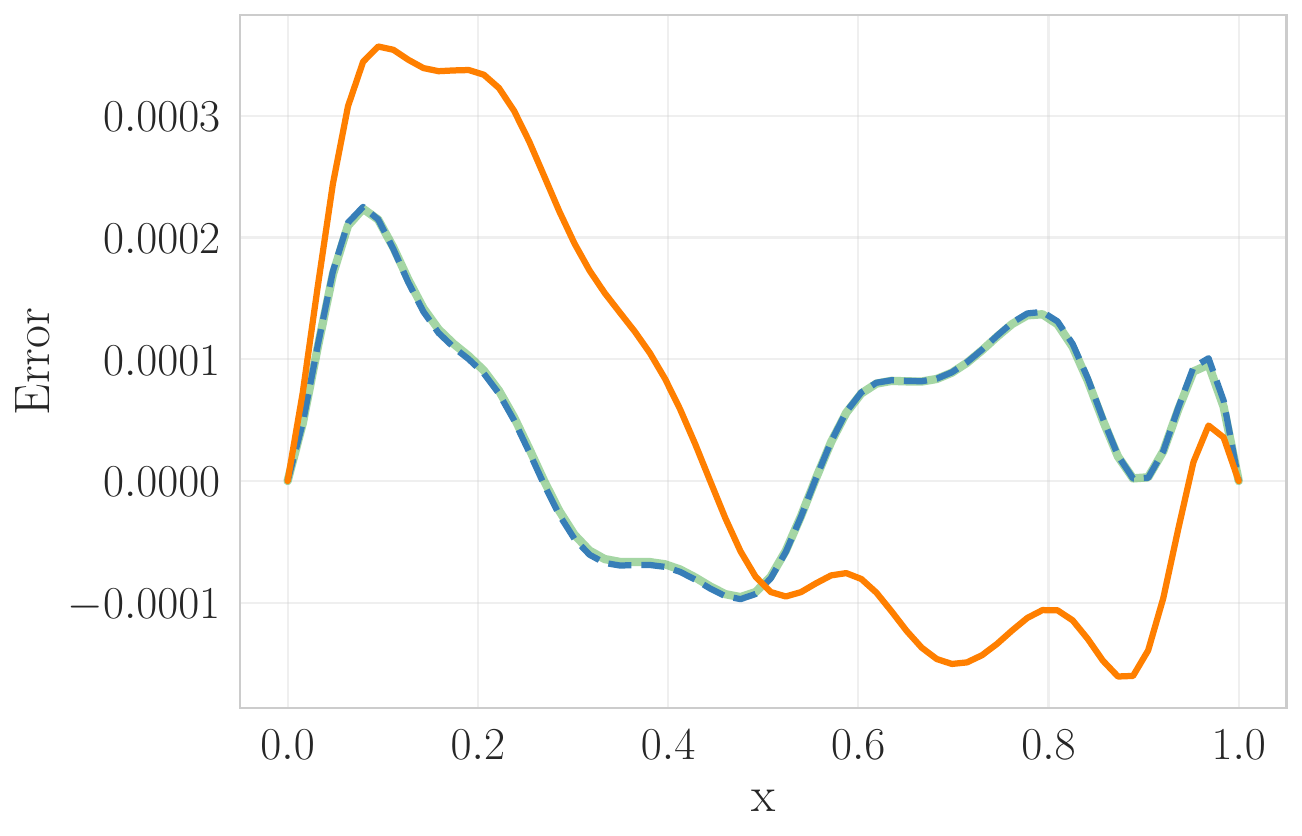}}\hfill
        \subcaptionbox{$t = 1.00$}{\includegraphics[width=0.48\linewidth]{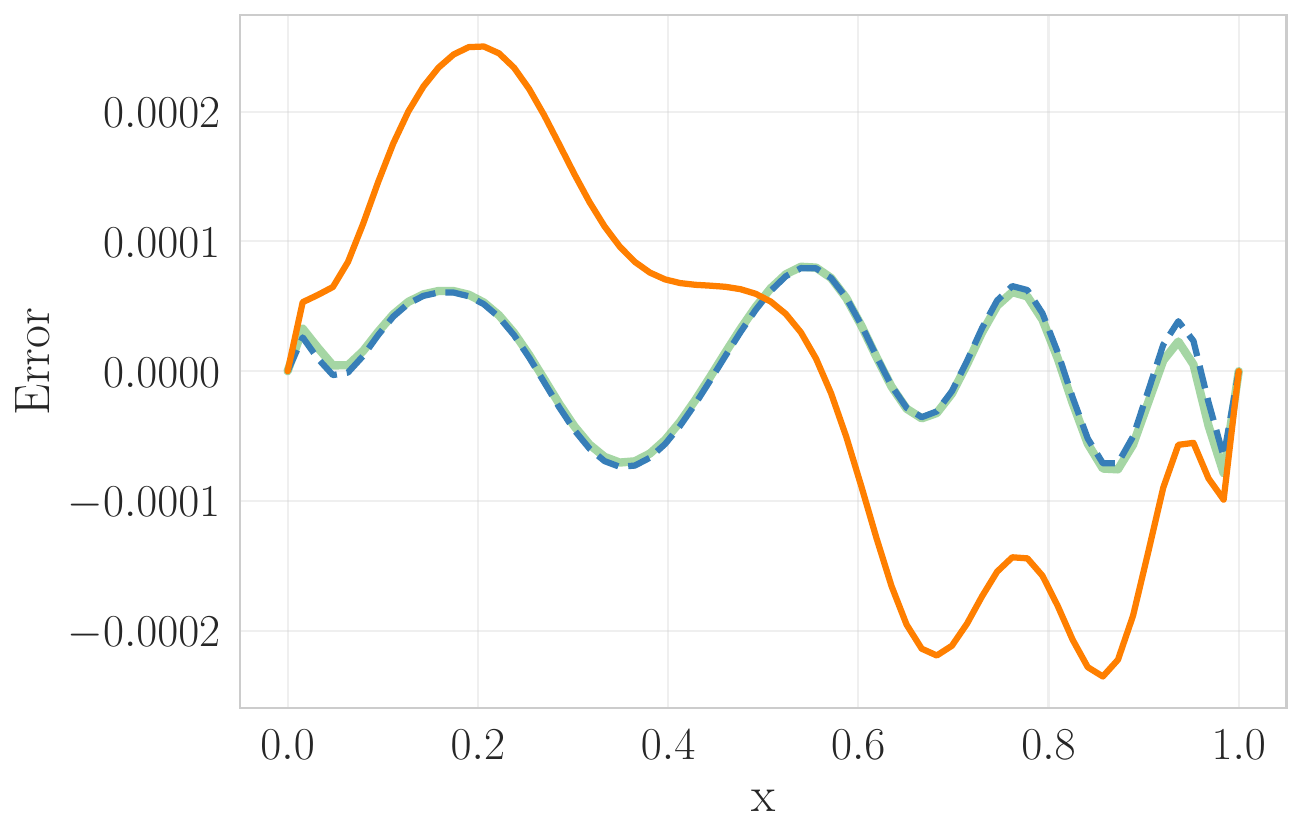}}
    \end{minipage}
    \hfill
    \begin{minipage}[c]{0.48\linewidth}
        \subcaptionbox{Error over time.}{\includegraphics[width=\linewidth]{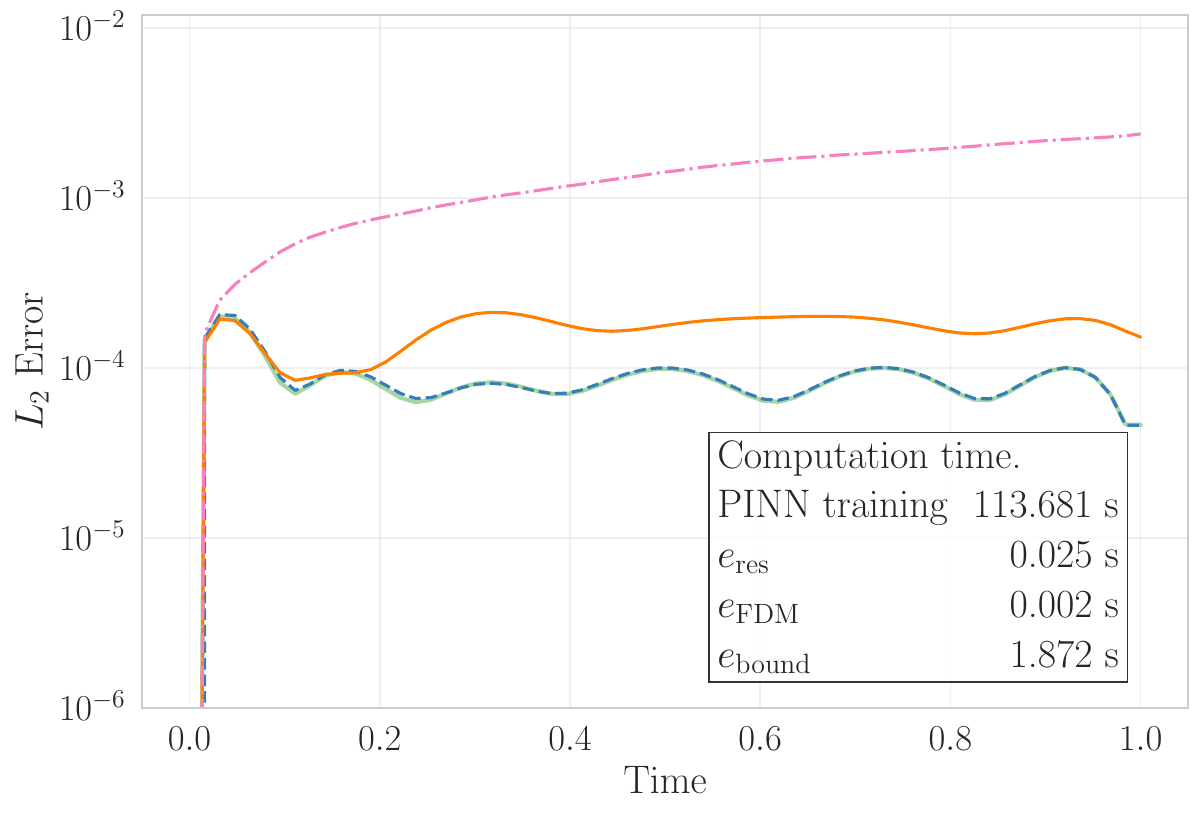}}
    \end{minipage}
    \caption{
    Error estimates for a \emph{well-trained} PINN on the heat equation, (a)--(d) at four different time-points across space and (e) $L_2$ error over time, including the bound $e_{\mathrm{bound}}$.
    The FDM-based methods use a $64 \times 64$ spatio-temporal grid (time slices shown at nearest grid point to specified values); $e_{\mathrm{bound}}$ uses $64$ spatial points with adaptive time integration (up to $165$ steps at $t=1$).
    }
    \label{fig:heat_error_comparison}
\end{figure}

\cref{fig:heat_error_comparison} presents results for the heat equation for a well-trained PINN.
At four time steps (\cref{fig:heat_error_comparison}a--d), the proposed residual-based estimate $e_{\mathrm{res}}$ tracks the true error closely in both shape and magnitude throughout the spatial domain.
The FDM reference $e_{\mathrm{FDM}}$, in contrast, performs worse at equal discretization. 
The error-over-time results (\cref{fig:heat_error_comparison}e) illustrate this gap from another perspective. 
Throughout time our method matches the true error closely and thus offers a complementary approach to the conservative error bound $e_{\mathrm{bound}}$, which overestimates the error.

\cref{fig:comparison_over_gridsizes} shows the accuracy in the error estimate as the $L_2$ error between the ground truth and the given method. 
For well-trained PINNs, $e_\mathrm{res}$ is consistently orders of magnitude more accurate than the FDM baseline $e_\mathrm{FDM}$. 
Notably, the error estimates are already accurate at low discretization.
For randomly initialized PINNs, $e_\mathrm{res}$ is less accurate than for well-trained ones, though it remains comparable to the baseline $e_\mathrm{FDM}$.
Notably, all configurations become more accurate with higher discretization, though at increased computational cost.

\newcommand{\compfigwidth}{0.18\linewidth}
\begin{figure}[b]
    \centering
    \fbox{\includegraphics[width=0.34\linewidth]{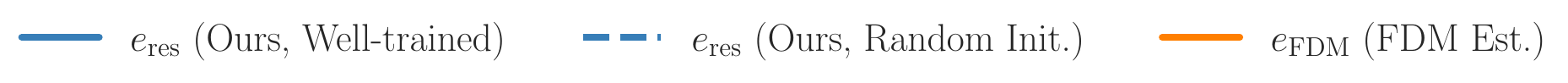}} \\ [0.3em]
    \begin{subfigure}{\compfigwidth}
        \includegraphics[width=\linewidth]{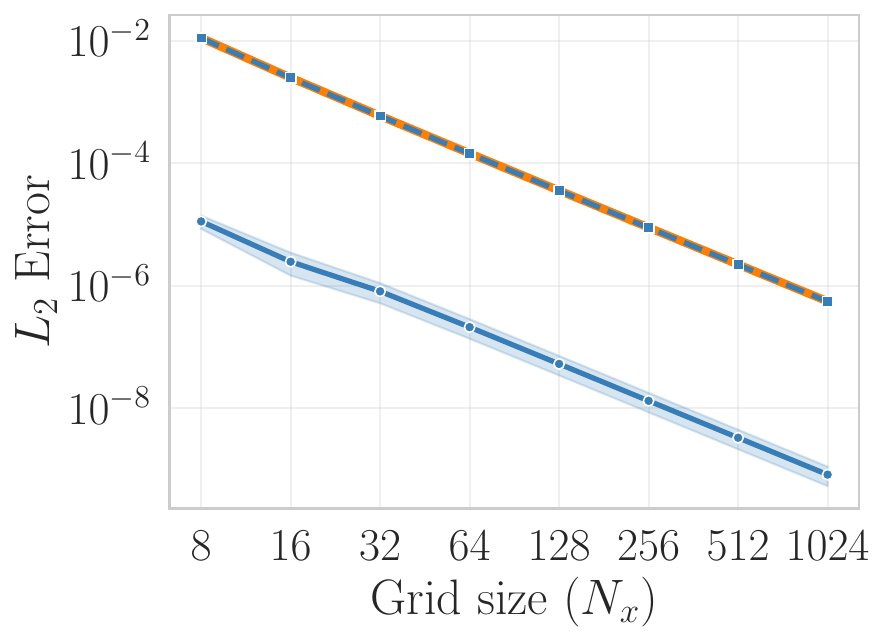}
        \caption{Poisson 1D}
    \end{subfigure}
    \begin{subfigure}{\compfigwidth}
        \includegraphics[width=\linewidth]{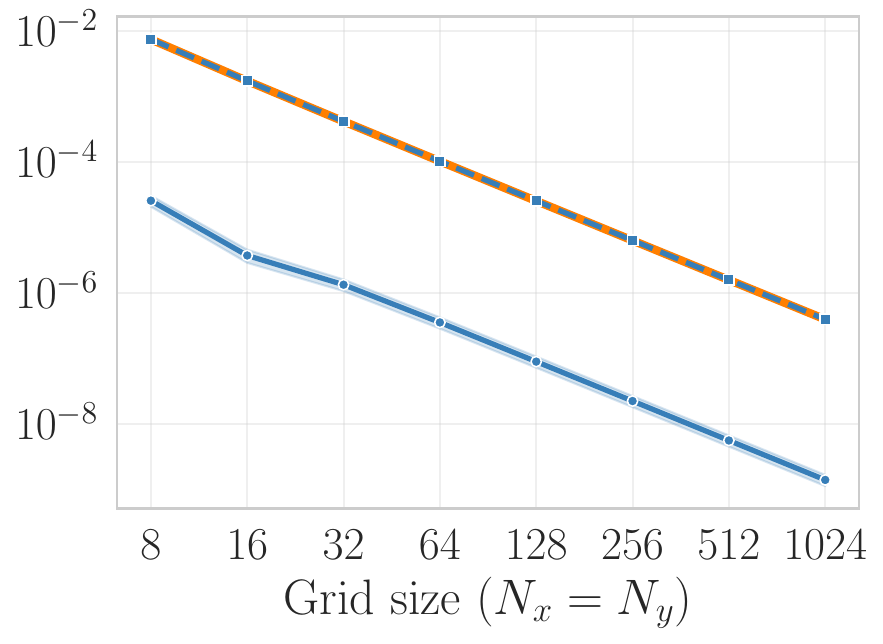}
        \caption{Poisson 2D}
    \end{subfigure}
    \begin{subfigure}{\compfigwidth}
        \includegraphics[width=\linewidth]{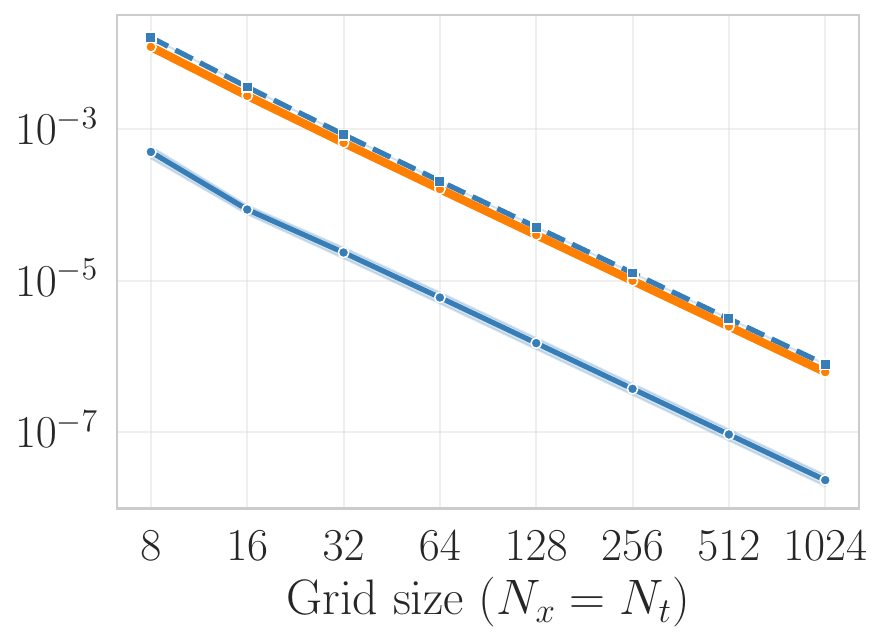}
        \caption{Heat}
    \end{subfigure}
    \begin{subfigure}{\compfigwidth}
        \includegraphics[width=\linewidth]{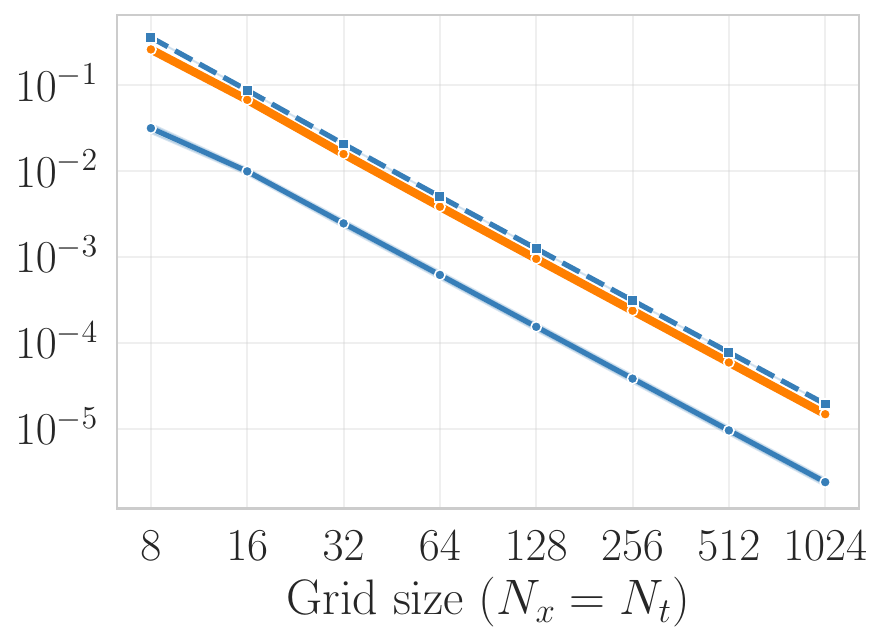}
        \caption{Drift-Diffusion}
    \end{subfigure}
    \begin{subfigure}{\compfigwidth}
        \includegraphics[width=\linewidth]{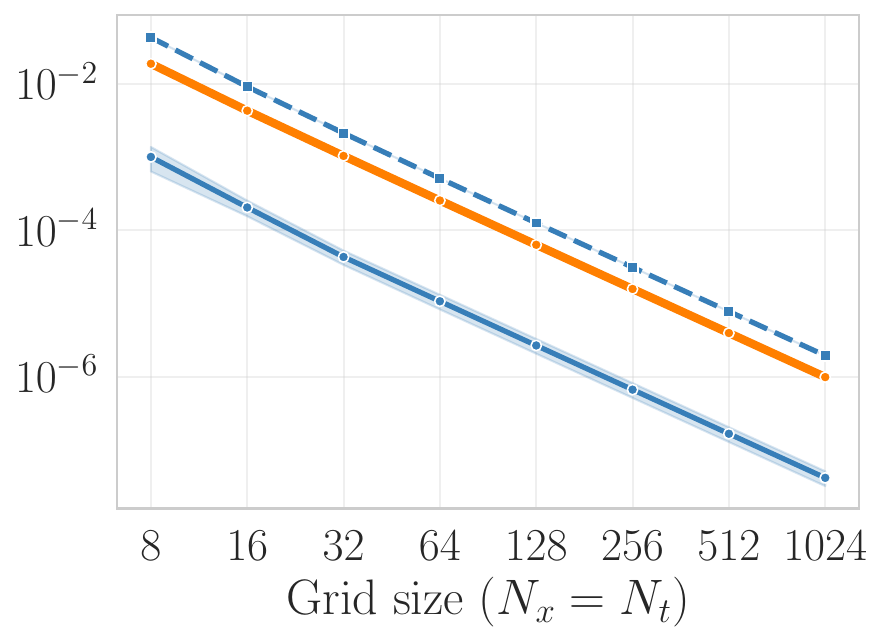}
        \caption{Wave}
    \end{subfigure}
    \caption{Accuracies of the error estimates across different discretizations, measured as $\norm{e_{\mathrm{true}} - e_{\mathrm{method}}}_2$ on respective grid points, for all benchmark problems. Averaged over 10 runs, bands show one standard deviation.}
    \label{fig:comparison_over_gridsizes}
\end{figure}

\paragraph{Discussion}
The experiments demonstrate the method's usefulness. Most notably, when the PINN is well-trained, solving the error equation using FDM achieves more accurate error estimates than through solving the original IBVP. 
We note, however, that the method is sensitive to the PINN's residual, achieving worse estimates for randomly initialized models. 
This sensitivity is likely tied to the smoothness of the PINN’s PDE residual, which affects the approximated derivatives used in the FDM calculation. Notably, our method also achieves accuracies comparable to the baseline for randomly initialized models.

Computationally, our method adds negligible overhead relative to PINN training (\cref{fig:heat_error_comparison}e), though the repeated backward passes required for residual evaluation render it more expensive than the baseline $e_{\mathrm{FDM}}$.
From an explainability perspective, the spatially resolved error maps produced by our method provide a diagnostic that localizes model failures. 
Our error maps reveal where and by how much a predicted approximation deviates from the true solution.
This enables local, informed decisions about whether to trust or reject PINN predictions at specific locations in the domain.

\paragraph{Limitations}
Our experiments are restricted to hard-constrained PINNs, where boundary and initial conditions are satisfied exactly.
We note that the output transformations, ensuring the hard-constraining, combined with normalized weight initialization, produce relatively smooth residuals even for untrained networks, which likely benefits our method.
Extending the framework to soft-constrained PINNs by including the errors at boundaries into the FDM computation is conceptually straightforward but
may, in practice, yield non-smooth residuals close to the boundaries. 
Whether the method retains similar robustness for soft-constrained PINNs remains an open question.

Furthermore, the considered benchmark problems are all linear and are constructed similarly. 
The practical transferability of the proposed approach to more complex problems, including non-linear, high-dimensional ones and those with complex geometries, 
may require more sophisticated solvers or alternative discretizations. 
From an interpretability point of view, the proposed method explains how far a prediction deviates but not why. 
Identifying sources of PINN failures remains future work. 
Finally, our method provides error estimates, not rigorous bounds.
The well-studied error theory of FDM schemes, however, opens up avenues to extend the proposed method to also include certified bounds for the approximated error.

\section{Conclusion}

Trustworthy deployment of PINNs requires explanations that go beyond training loss and reveal where and by how much predictions deviate from the true solution.
We present a lightweight post-hoc method for estimating the prediction error of PINNs by solving the associated error equation using finite difference methods. 
This approach requires no knowledge of the true solution and yields accurate, spatially resolved error maps that serve as quantitative diagnostics of PINN prediction quality, identifying where a model can be trusted and where it fails. 
Across all benchmark problems, our method consistently achieves more accurate error estimates than a direct FDM solution for well-trained and hard-constrained PINNs, and remains competitive for randomly initialized ones. 
Promising extensions include more complex problem classes, as well as building on top of the well-understood error theory of FDM schemes towards certified bounds derived from the estimated error.

\begin{acknowledgments}
  This work was supported by the Fraunhofer Internal Programs under Grant No. PREPARE 40-08394.
\end{acknowledgments}
\section*{Declaration on Generative AI}
  

 During the preparation of this work, the authors used Claude Sonnet 4.5, 4.6, Claude Opus 4.5, 4.6 in order to: check spelling and grammar, improve writing style, and as a coding assistant.

\bibliography{references_nodoi_nourl}


\end{document}

%% file: math_commands.tex
\usepackage{amsmath,amsfonts,bm}

\def\eqref#1{equation~\ref{#1}}

\def\1{\bm{1}}

\DeclareMathAlphabet{\mathsfit}{\encodingdefault}{\sfdefault}{m}{sl}
\SetMathAlphabet{\mathsfit}{bold}{\encodingdefault}{\sfdefault}{bx}{n}



%% file: commands.tex
\usepackage{soul}

\newcommand{\norm}[1]{\left\lVert#1\right\rVert}

\newcommand\blfootnote[1]{%
  \begingroup
  \renewcommand\thefootnote{}\footnote{#1}%
  \addtocounter{footnote}{-1}%
  \endgroup
}